\documentclass[invited]{ieice}

\usepackage{graphicx}
\usepackage{amsmath}
\usepackage{newtxtext}
\usepackage[varg]{newtxmath}
\usepackage{algorithmic}
\usepackage{algorithm}
\usepackage{cite}
\usepackage{color}

\field{}
\title{Image and Model Transformation with Secret Key for Vision Transformer}
\authorlist{
 \authorentry[kiya@tmu.ac.jp]{Hitoshi Kiya}{f}{labelA}\MembershipNumber{}
 \authorentry[iijima-ryota@ed.tmu.ac.jp]{Ryota Iijima}{n}{labelA}\MembershipNumber{}
 \authorentry[april-pyone-maung-maung@ed.tmu.ac.jp]{MaungMaung AprilPyone}{n}{labelA}\MembershipNumber{}
 \authorentry[ykinoshita@tmu.ac.jp]{Yuma Kinoshita}{m}{labelA}
}
\affiliate[labelA]{The author is with Tokyo Metropolitan University, Hino, 191-0065, Japan.}
\affiliate[affiliate label]{The author is with the 
}

\received{2015}{1}{1}
\revised{2015}{1}{1}

\begin{document}
\maketitle
\begin{summary}
In this paper, we propose a combined use of transformed images and vision transformer (ViT) models transformed with a secret key.
We show for the first time that models trained with plain images can be directly transformed to
models trained with encrypted images on the basis of the ViT architecture, and the performance of the transformed models is the same as models trained with plain images when using test images encrypted with the key.
In addition, the proposed scheme does not require any specially prepared data for training models or network modification, so it also allows us to easily update the secret key.
In an experiment, the effectiveness of the proposed scheme is evaluated in terms of performance degradation and model protection performance in an image classification task on the CIFAR-10 dataset.
\end{summary}

\section{Introduction}
Machine learning (ML) algorithms have been deployed in many applications including security-critical ones such as biometric authentication,  automated driving , and medical image analysis \cite{lecun2015deep, liu2019recent}.
However, training successful models requires three ingredients: a huge amount of data, GPU accelerated computing resources, and efficient algorithms, and it is not a trivial task.
In fact, collecting images and labeling them is also costly and will also consume a massive amount of resources.
Therefore, trained ML models have great business value.
Considering the expenses necessary for the expertise, money, and time taken to train a model, a model should be regarded as a kind of intellectual property (IP).
In addition, generally, data contains sensitive information and it is difficult to train a model while preserving privacy.
In particular, data with sensitive information cannot be transferred to untrusted third-party cloud environments (cloud GPUs and TPUs) even though they provide a powerful computing environment \cite{huang2014survey, moo2013p3, lagendijk2013encrypted, fredrikson2015model, shokri2017membership, christian2014intriguing, siva2020adversarial}.
Accordingly, it has been challenging to train/test a ML model with encrypted images as one way for solving these issues \cite{kiya2022overview}. \par
Learnable image encryption, which has given new solutions to the above issues, is encryption that allows us not only to generate visually protected images to protect personally identifiable information included in an image, such as an individual or the time and location of the taken photograph, but to also apply encrypted images to a ML algorithm in the encrypted domain \cite{kiya2022overview}.
In addition, image encryption with a secret key, referred to as image transformation with a secret key, can embed unique features controlled with the key into images.
The use of the unique features was demonstrated to be effective in applications such as  adversarial defense  and model protection \cite{aprilpyone2021block, maungmaung2021a, maung2020encryption, maung2021ensemble, april2022privacy, qi2022privacy}.
However, even though many image transformation methods with a secret key have been studied so far for application to such applications, no conventional methods can avoid the influence of image transformation.
In other words, the use of transformed images degrades the performance of models compared with models trained with plain images \cite{aprilpyone2021block, maungmaung2021a, maung2020encryption, maung2021ensemble, april2022privacy, qi2022privacy}.
In addition, if we want to update the key, models have to be re-trained by using a new key. \par

In this paper, we show that the use of the vision transformer (ViT) \cite{Alexey2021an} allows us to reduce the influence of block-wise encryption thanks to its architecture.
After that, to overcome the problems with conventional image transformation, we propose a novel framework for ML algorithms with encrypted images that uses ViT.

In the framework, a model trained with plain images is also transformed with a secret key using the unique features in addition to test images, and the combined use of the transformed model and test images is proposed for using ML algorithms in the encrypted domain.
The proposed scheme allows us not only to obtain the same performance as models trained with plain images but to also update the secret key easily.
In an experiment, the effectiveness of the proposed scheme is evaluated in terms of performance degradation and model protection performance in an image classification task on the CIFAR-10 dataset.

\section{Related Work}
Conventional image transformation for machine learning and ViT are summarized here.

\subsection{Image Transformation for Machine Learning}
Various image transformation methods with a secret key, often referred to as perceptual image encryption or image cryptography, have been studied so far for many applications.
Perceptional encryption can offer encrypted images that are described as bitmap images, so the encrypted images can be directly applied to image processing algorithms.
In addition, encrypted images can be decrypted even when noise is added to them, although  the use of standard encryption algorithms such as DES and AES  cannot. \par
Figure 1 shows typical applications of image transformation with a key.
Image transformation with a key allows us not only to protect visual information on plain images but to also embed unique features controlled with the key into images.
The use of visually protected images has enabled various kinds of applications.
One of the origins of image transformation with a key is in block-wise image encryption schemes for encryption-then-compression (EtC) systems \cite{chuman2019encryption, chuman2017security, zhou2014designing, ghonge2014a, liu2018ecg, liu2010efficient, hu2014a, johnson2004on, methaq2016an, watanabe2015an}.
Image encryption prior to image compression is required in certain practical scenarios such as secure image transmission through an untrusted channel provider.
An EtC system is used in such scenarios, although the traditional way of securely transmitting images is to use a compression-then-encryption (CtE) system.
Compressible encryption methods have been applied to privacy-preserving compression, data hiding, and image retrieval\cite{Shoko2020a, iida2020privacy, iida2019an} in cloud environments.
In addition, visually protected images have been demonstrated to be effective in fake image detection \cite{liu2022block} and various learning algorithms \cite{kiya2022overview, kawamura2020aprivacy, bandoh2020distributed, takayuki2020secure, nakachi2020privacy, ibuki2016unitary, maekawa2019privacy}. \par
In this paper, we focus on image transformation methods for machine learning including deep neural networks (DNNs), called learnable encryption.
Learnable encryption enables us to directly apply encrypted data to a model as training and testing data.
Encrypted images have no visual information on plain images in general, so privacy-preserving learning can be carried out by using visually protected images.
In addition, the use of a secret key allows us to embed unique features controlled with the key into images.
Adversarial defense \cite{aprilpyone2021block, maung2020encryption, maung2021ensemble}, access control \cite{chen2018protect, chen2019deepattest, maungmaung2021a}, and DNN watermarking \cite{maung2021ensemble, uchida2017embedding, chen2019deepmarks, rouhani2018deepsigns, fan2021deepip, adi2018turning, zhang2018protecting, sakazawa2019visual, le2019adversarial} are carried out with encrypted data using the unique features. \par

\begin{figure*}[t]
    \centering
    \includegraphics[scale=0.25]{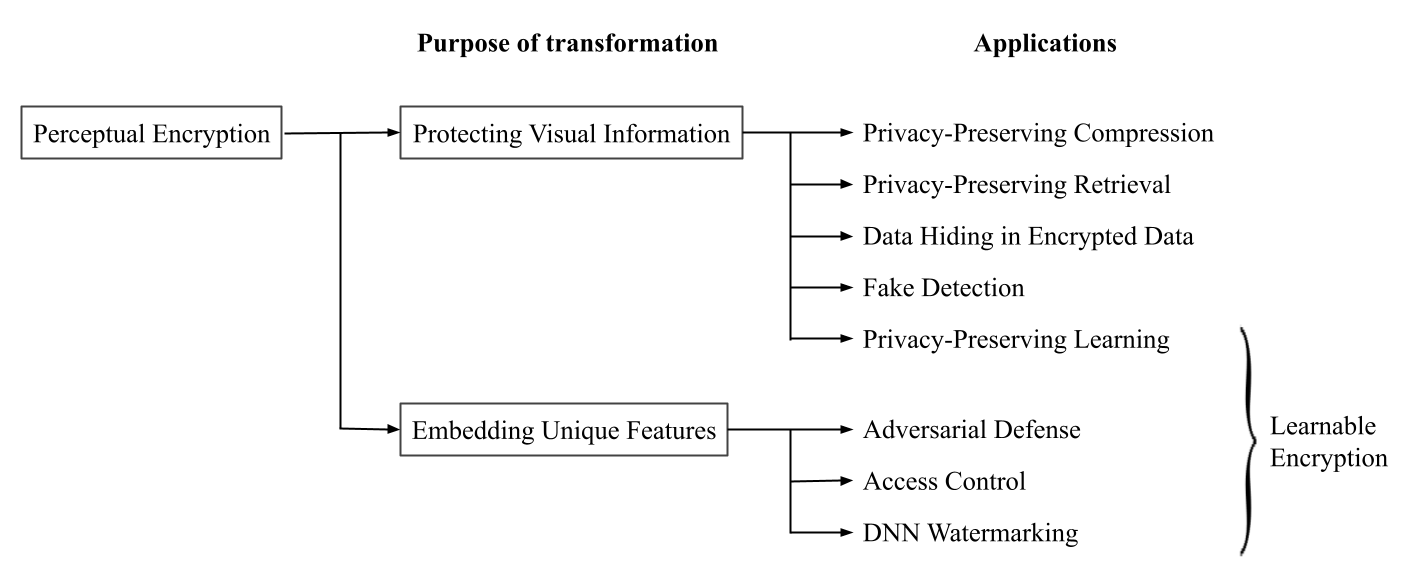}
    \caption{Applications of perceptual image encryption}
    \label{fig1}
\end{figure*}

\subsection{Vision Transformer}
The transformer architecture has been widely used in natural language processing (NLP) tasks \cite{devlin2019bert}.
The vision transformer (ViT) \cite{Alexey2021an} has also provided excellent results compared with state-of-the-art convolutional networks.
Following the success of ViT, a number of isotropic networks (with the same depth and resolution across different layers in the network) have been proposed such as MLP-Mixer \cite{ilya2021mlpmixer}, ResMLP \cite{hugo2021resmlp}, CycleMLP \cite{chen2022cyclemlp}, gMLP \cite{liu2021pay}, vision permutator \cite{Hou2022VisionPA}, and ConvMixer \cite{trockman2022patches}.
In this paper, we focus on ViT because it utilizes patch embedding and position embedding (see Fig. 2).
Figure 2 illustrates the architecture of ViT.
The main procedure of ViT is given as follows:

\begin{enumerate}
    \item Split an image into fixed-size patches, and linearly embed each of them.
    \item Add position embedding to patch embedding.
    \item Feed the resulting sequence of vectors to a standard transformer encoder.
    \item Feed the output of the transformer to a multi-layer perceptron (MLP), and get a result.
\end{enumerate}

\noindent
ViT utilizes patch embedding and position embedding.
In this paper, 
we point out that the embedding structure enables us to reduce the influence of block-wise encryption.
In patch embedding, patches are mapped to vectors, and in position embedding, the position information is embedded.
If every patch (block) is transformed with the same key, pixel shuffling and bit flipping can be express as the operation of patch embedding.
In addition, block permutation can be given as the operation of position embedding.
The relation between block-wise encryption and the embedding structure is demonstrated to avoid the influence of block-wise image encryption.

\begin{figure}[t]
    \centering
    \includegraphics[scale=0.18]{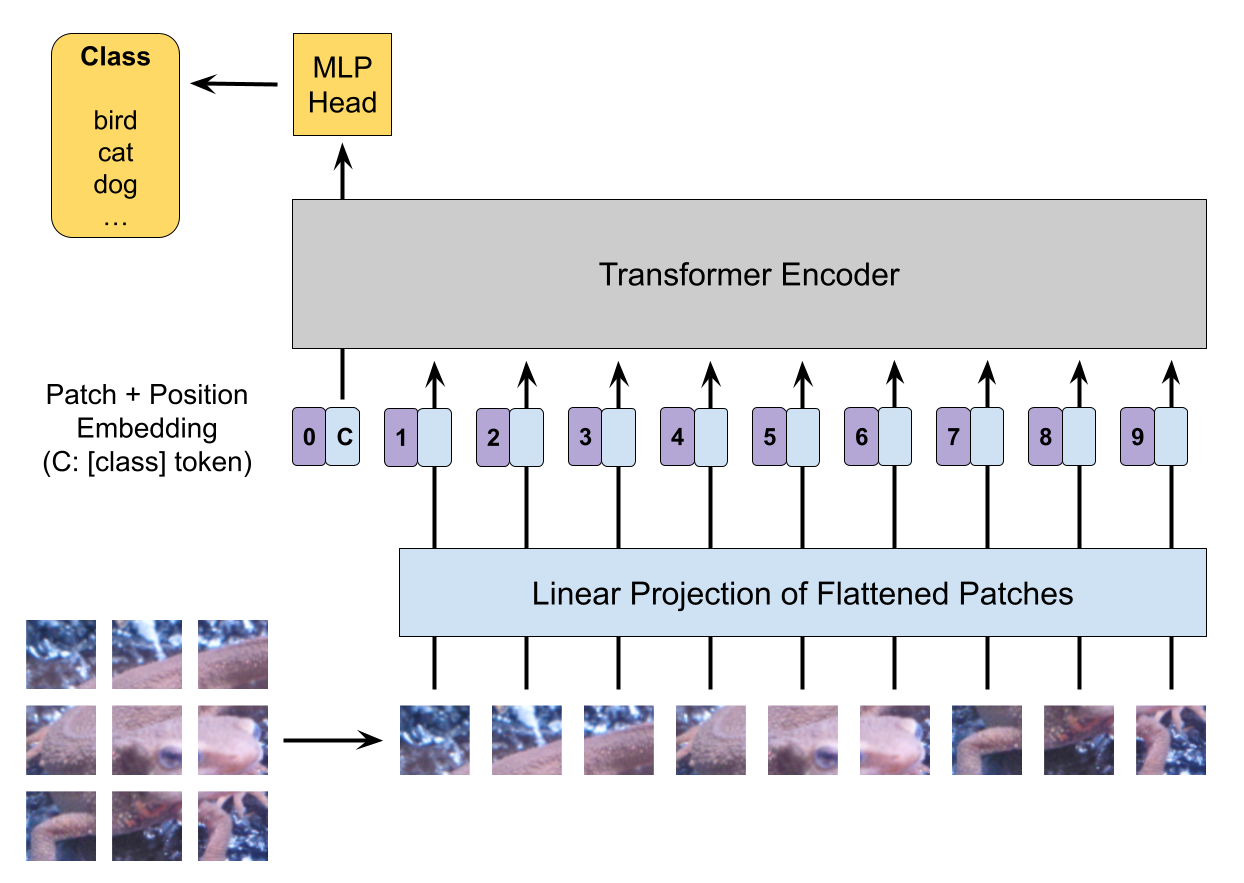}
    \caption{ Architecture of ViT \cite{Alexey2021an} }
    \vspace{-10mm}
    \label{fig2}
\end{figure}

\section{Image and Model Transformation with Secret Key}
An image transformation method with a secret key is proposed here.
The method makes it possible to simultaneously use both transformed images and models.

\subsection{Notation}
The following notations are utilized throughout this paper.

\begin{itemize}
    \item $W$, $H$, and $C$ denote the width, height, and the number of channels of an image, respectively.
    \item The tensor $x \in [0, 1]^{C \times W \times H}$ represents an input color image.
    \item The tensor $x' \in [0, 1]^{C \times W \times H}$ represents an encrypted image.
    \item $M$ is the block size of an image.
    \item Tensors $x_\mathrm{b}$ and ${x'}_\mathrm{b} \in [0, 1]^{W_\mathrm{b} \times H_\mathrm{b} \times p_\mathrm{b}}$ are a block image and an encrypted block image, respectively, where $W_\mathrm{b} = \frac{W}{M}$ is the number of blocks across width $W$, $H_\mathrm{b} = \frac{H}{M}$ is the number of blocks across height $H$, and $p_\mathrm{b} = M \times M \times C$ is the number of pixels in a block.
    
    We assume that $W$ and $H$ are divisible by $M$, so $W_\mathrm{b}$ and $H_\mathrm{b}$ are positive integers.
    
    \item A pixel value in a block image ($x_\mathrm{b}$ or ${x'}_\mathrm{b}$) is denoted by $x_\mathrm{b}(w, h, c)$ or ${x'}_\mathrm{b}(w, h, c)$, where $w \in \{ 0, ..., W_\mathrm{b} - 1 \}$, $h \in \{ 0, ..., H_\mathrm{b} - 1 \}$, and $c \in \{ 0, ..., p_\mathrm{b} - 1 \}$ are indices corresponding to the dimension of $x_\mathrm{b}$ of ${x'}_\mathrm{b}$.
    \item $x_\mathrm{b}(w, h, :)$ denotes a vector $(x_\mathrm{b}(w, h, 0), ..., x_\mathrm{b}(w, h, p_\mathrm{b} - 1))$ of a tensor $x_\mathrm{b}$.
    \item $x_\mathrm{b}(:, :, c)$ denotes a matrix of a tensor $x_\mathrm{b}$ as given in
    
    \begin{equation}
        \begin{pmatrix} 
            x_\mathrm{b}(0, 0, c) & \dots & x_\mathrm{b}(W_\mathrm{b} - 1, 0, c) \\
            \vdots & \ddots & \vdots \\
            x_\mathrm{b}(0, H_\mathrm{b} - 1, c) & \dots  & x_\mathrm{b}(W_\mathrm{b} - 1, H_\mathrm{b} - 1, c)
        \end{pmatrix}.
    \end{equation}
    
    \item $B$ is a block in an image , and its dimension is $M \times M \times C$.
    \item $\hat{B}$ is a flattened version of block $B$, and its dimension is $1 \times 1 \times p_\mathrm{b}$.
    \item $P$ is the patch size of an image.
\end{itemize}

\subsection{Overview}
Figure 3 shows the scenario of the proposed scheme, where it is assumed that the  classification model builder  is trusted, and the  classification service provider  is untrusted.
The  classification model builder  trains a model by using plain images and transforms the trained model with a secret key  where the transformation by using secret keys is performed only on the prepossessing part of the transformer encoder. 
The transformed model is given to the  classification service provider , and the key is sent to a client.
The client prepares a transformed test image with the key and sends it to the provider.
The provider applies it to the transformed model to obtain a classification result, and the result is sent back to the client. \par
Note that the provider has neither a key nor plain images.
The proposed scheme enables us to achieve this scenario without any performance degradation compared with the use of plain images.

\begin{figure*}[t]
    \centering
    \includegraphics[scale=0.35]{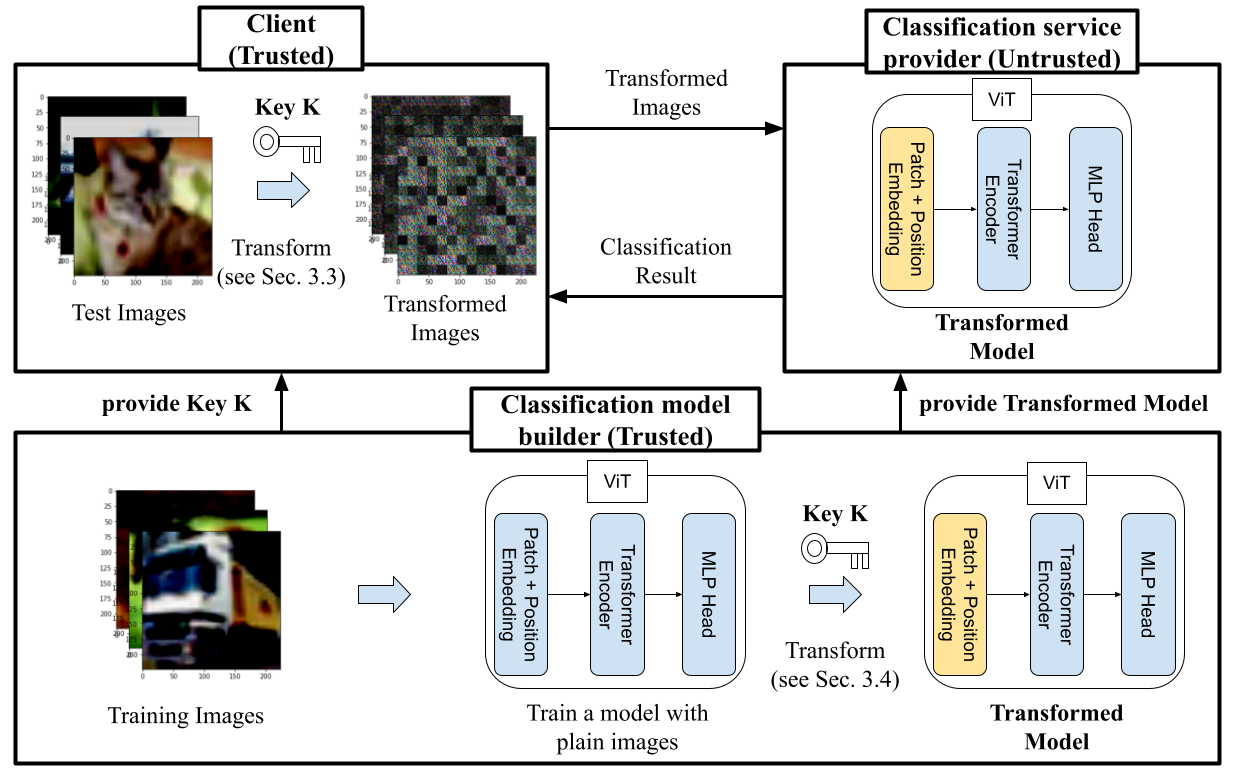}
    \caption{ Scenario of proposed scheme }
    \label{fig3}
\end{figure*}

\begin{figure*}[t]
    \centering
    \includegraphics[scale=0.34]{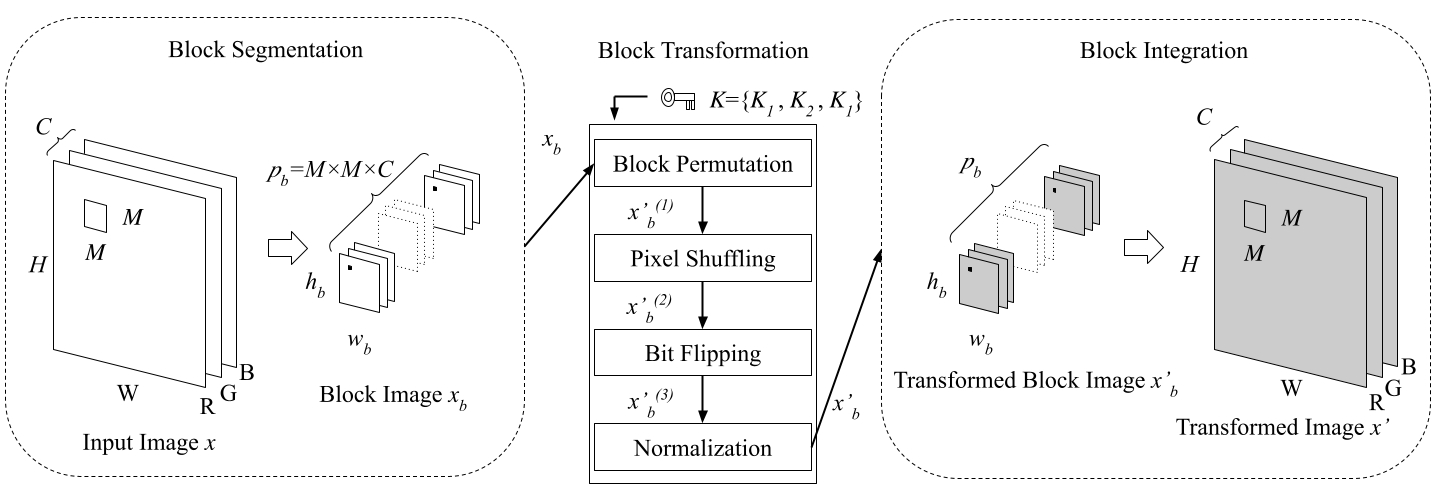}
    \caption{Procedure of block-wise transformation}
    \label{fig4}
\end{figure*}

\subsection{Image Transformation}
A block-wise image transformation with a secret key is proposed for application to test images.
As shown in Fig. 4, the procedure of the transformation consists of three steps: block segmentation, block transformation, and block integration. 
To transform an image $x$, we first divide $x$ into $W_\mathrm{b} \times H_\mathrm{b}$ blocks, as in $\left \{ B_{1 1},B_{1 2},...,B_{W_\mathrm{b} H_\mathrm{b}} \right \}$.
In this paper, we assume that the block size of the segmentation is the same as the patch size of ViT.
Next, each block is flattened, and it is concatenated again to obtain a block image $x_\mathrm{b}$.
Then, $x_\mathrm{b}$ is transformed to ${x'}_\mathrm{b}$ in accordance with block transformation with key $K$.
Finally, ${x'}_\mathrm{b}$ is transformed so that it has the same $C \times H \times W$ dimensions as those of the original image, and encrypted image $x'$ is obtained. \par
In addition, the block transformation is carried out by using the four operations shown in Fig. 4.
Details on each operation are given below.

\renewcommand{\thesubsubsection}{\Alph{subsubsection}}
\subsubsection{Block Permutation}
\begin{enumerate}
    \item Generate a random permutation vector $\boldsymbol{v_{\mathrm{A}}} = (v_0, v_1, ..., v_k, ..., v_{k'}, ..., v_{W_{\mathrm{b}} \times H_{\mathrm{b}} -1})$ that consists of randomly permuted integers from $0$ to $W_{\mathrm{b}} \times H_n - 1$ by using a key $K_1$, where $k, k' \in \{0, ..., H_{\mathrm{b}} \times W_{\mathrm{b}} - 1 \}$, and $v_k \neq v_{k'}$ if $k \neq k'$.
    \item Blocks are permutated to replace $x_\mathrm{b}$ with $x'^{(1)}_{\mathrm{b}}$ by using vector $\boldsymbol{v_{\mathrm{A}}}$ (see Algorithm 1).
\end{enumerate}

\begin{algorithm}[H]
    \caption{Block permutation}
    \begin{algorithmic}
    \renewcommand{\algorithmicrequire}{\textbf{Input:}}
    \renewcommand{\algorithmicensure}{\textbf{Output:}}
    \REQUIRE $x_{\mathrm{b}}, K_1$
    \ENSURE $x'^{(1)}_{\mathrm{b}}$
    \STATE {Generate a vector} $\boldsymbol{v_{\mathrm{A}}}$ {using key} $K_1$ {.}
    \STATE $y_{\mathrm{b}} \leftarrow (x_{\mathrm{b}}[0,0,:], x_{\mathrm{b}}[1,0,:],...,x_{\mathrm{b}}[W_{\mathrm{b}}-1,H_{\mathrm{b}}-1,:])$
    \STATE $i \leftarrow 0$
    \WHILE{$i < H_\mathrm{b} \times W_\mathrm{b}$}
        \STATE $y'_{\mathrm{b}}[i] \leftarrow y_{\mathrm{b}}[\boldsymbol{v_{\mathrm{A}}}[i]]$
        \STATE $i \leftarrow i + 1$
    \ENDWHILE
    \STATE $w \leftarrow 0$
    \STATE $h \leftarrow 0$
    \WHILE{$h < H_{\mathrm{b}}$}
        \WHILE{$w < W_{\mathrm{b}}$}
            \STATE $x'^{(1)}_{\mathrm{b}}[w,h,:] \leftarrow y'_{\mathrm{b}}[w + h \times W_{\mathrm{b}} ]$
            \STATE $w \leftarrow w + 1$
        \ENDWHILE
        \STATE $h \leftarrow h + 1$
    \ENDWHILE
    \RETURN $x'^{(1)}_{\mathrm{b}}$
    \end{algorithmic}
\end{algorithm}

\subsubsection{Pixel Shuffling}
\begin{enumerate}
    \item Generate a random permutation vector $\boldsymbol{v_{\mathrm{B}}} = (v_0, v_1, ..., v_k, ..., v_{k'}, ..., v_{p_{\mathrm{b}} -1})$ by using a key $K_2$, where $k, k' \in \{0, ..., p_{\mathrm{b}} - 1 \}$, and $v_k \neq v_{k'}$ if $k \neq k'$.
    \item Pixels in each block are shuffled by vector $\boldsymbol{v_{\mathrm{B}}}$ as (see Algorithm 2).
    
    \begin{equation}
        x'^{(2)}_{\mathrm{b}}(w, h, k) = x'^{(1)}_{\mathrm{b}}(w, h, v_k).
    \end{equation}
\end{enumerate}

\begin{algorithm}[H]
    \caption{Pixel shuffling}
    \begin{algorithmic}
    \renewcommand{\algorithmicrequire}{\textbf{Input:}}
    \renewcommand{\algorithmicensure}{\textbf{Output:}}
    \REQUIRE {$x'^{(1)}_{\mathrm{b}}, K_2$ }
    \ENSURE $x'^{(2)}_{\mathrm{b}}$
    \STATE {Generate a vector} $\boldsymbol{v_{\mathrm{B}}}$ {using key} $K_2$ {.}
    \STATE $i \leftarrow 0$
    \WHILE{$i < p_\mathrm{b}$}
        \STATE $x'^{(2)}_{\mathrm{b}}[:, :, i] \leftarrow x'^{(1)}_{\mathrm{b}}[:,:,\boldsymbol{v_{\mathrm{B}}}[i]]$
        \STATE $i \leftarrow i + 1$
    \ENDWHILE
    \RETURN $x'^{(2)}_{\mathrm{b}}$
    \end{algorithmic}
\end{algorithm}

\subsubsection{Bit Flipping}
\begin{enumerate}
    \item Generate a random binary vector $\boldsymbol{r} = (r_0,...,r_k,...,r_{p_{\mathrm{b}} - 1})$, $r_k \in \{ 0,1 \}$ by using a key $K_3$. To keep the transformation consistent, $r$ is distributed with $50\%$ of ``0''s and $50\%$ of ``1''s.
    \item Apply negative-positive transformation on the basis of $\boldsymbol{r}$ as
    
    \begin{equation}
        x'^{(3)}_{\mathrm{b}}(w, h, k) = \left \{
        \begin{aligned}
        & x'^{(2)}_{\mathrm{b}}(w,h,k) && (r_k=0) \\
        & 1 - x'^{(2)}_{\mathrm{b}}(w,h,k) &&(r_k=1).
        \end{aligned}
    \right .
    \end{equation}
    
\end{enumerate}

\begin{algorithm}[H]
    \caption{Bit flipping}
    \begin{algorithmic}
    \renewcommand{\algorithmicrequire}{\textbf{Input:}}
    \renewcommand{\algorithmicensure}{\textbf{Output:}}
    \REQUIRE $x'^{(2)}_{\mathrm{b}}, K_3$
    \ENSURE $x'^{(3)}_{\mathrm{b}}$
    \STATE {Generate a vector} $\boldsymbol{r}$ {using key} $K_3$ {.}
    \STATE $i \leftarrow 0$
    \WHILE{$i < p_\mathrm{b}$}
        \IF{$\boldsymbol{r}[i] = 0$}
            \STATE $x'^{(3)}_{\mathrm{b}}[:,:,i] \leftarrow  x'^{(2)}_{\mathrm{b}}[:,:,i]$
        \ELSE
            \STATE  $x'^{(3)}_{\mathrm{b}}[:,:,i] \leftarrow 1 - x'^{(2)}_{\mathrm{b}}[:,:,i]$ 
        \ENDIF
        \STATE $i \leftarrow i + 1$
    \ENDWHILE
    \RETURN $x'^{(3)}_{\mathrm{b}}$
    \end{algorithmic}
\end{algorithm}

\subsubsection{Normalization}
Various normalization methods are widely used to improve the training stability, optimization efficiency, and generalization ability of DNNs.
In this paper, we use a normalization method to achieve the combined use of transformed images and models. \par

From Eq. (3), if $r_k=0$, a pixel ${x'}^{(3)}_\mathrm{b}(w, h, k)$ is replaced with ${x'}_\mathrm{b}(w, h, k)$ as
\begin{equation}
    \begin{aligned}
    {x'}_\mathrm{b}(w, h, k) &= \frac{{x'}^{(3)}_\mathrm{b}(w, h, k) - 1/2}{1/2} \\
    &= \frac{{x'}^{(2)}_\mathrm{b}(w, h, k) - 1/2}{1/2}. \\
    \end{aligned}
\end{equation}
In contrast, if $r_k=1$, a pixel ${x'}^{(3)}_\mathrm{b}(w, h, k)$ is replaced as follows.
\begin{equation}
    \begin{aligned}
    {x'}_\mathrm{b}(w, h, c) &= \frac{{x'}^{(3)}_\mathrm{b}(w, h, c) - 1/2}{1/2} \\
    &= 2{x'}^{(3)}_\mathrm{b}(w, h, c) - 1 \\
    &= 2(1 - {x'}^{(2)}_\mathrm{b}(w, h, c)) - 1 \\
    &= 1 - 2{x'}^{(2)}_\mathrm{b}(w, h, c) \\
    &= - \frac{{x'}^{(2)}_\mathrm{b}(w, h, c) - 1/2}{1/2} \\
    \end{aligned}
\end{equation}

Therefore, bit flipping with normalization can be regarded as an operation that reverses the positive or negative sign of a pixel value.
This property allows us to use the model encryption that will be described later. \par

The above encryption steps are the same as those of a number of conventional methods \cite{kiya2022overview, aprilpyone2021block}, but the performance of conventional models is degraded due to the influence of encryption when the encryption steps are used in the conventional schemes.
In contrast, the proposed method is demonstrated to avoid the influence of encryption in this paper, which is one of the reasons to apply the algorithm written in each step.
In addition, the encryption steps can be expressed as a linear transform as described in the paper.
Other encryption steps can also be used under the framework of the proposed scheme, if they are expressed as a linear transform such as a random matrix.

\subsection{Model Transformation}
In model transformation, some parameters in models trained with plain images are transformed by using a secret key.
In this paper, a model transformation method is proposed that can achieve the combined use of models and images transformed with the same key. \par
ViT utilizes patch embedding and position embedding (see Fig. 2), so it has the following two properties.

\begin{enumerate}
    \item Patch-order invariance of transformer encoder: the output of the transformer encoder corresponding to an input patch is independent of the order of input patches.
    \item Ability to adapt to pixel order by patch embedding: patch embedding can be adapted to pixel shuffling and bit flipping because they can be expressed as an invertible linear transformation as described below.
\end{enumerate}

In the proposed scheme, it is assumed that the patch size $P$ used for patching is the same as the block size used for image encryption, and the number of patches is equal to that of blocks in an image.
The transformation of parameters in trained models is described below.

\subsubsection{Position Embedding and Patch Embedding}
In ViT \cite{Alexey2021an}, all segmented patches are flattened.
A pixel value in the flattened patches is given by a pixel value in the original image as

\begin{equation}
    \begin{aligned}
    & \boldsymbol{x}^i_{\mathrm{p}}[k] = x[h, w, c], \\
    & h = \left \lfloor \frac{i-1}{W/P} \right \rfloor P + \left \lfloor \frac{k-1 \bmod P^2}{P} \right \rfloor, \\
    & w = \left ( 
    (i-1) \bmod (W/P) \right ) P + ((k-1) \bmod P), \\
    & c = \left \lfloor \frac{k-1}{P^2} \right \rfloor, \\
    & \boldsymbol{x}^i_{\mathrm{p}} \in \mathbb{R}^{P^2 C}, i \in \{ 1,2,...,N \}, k \in \{ 1,2,...,P^2 C \}
    \end{aligned}
\end{equation}

\noindent
where  $\boldsymbol{x}^i_{\mathrm{p}}[k]$ is a pixel value in the $i$-th patch, and  $N=HW/P^2$ is the number of patches.
To simplify the discussion, we assume that $H$ and $W$ are divisible by $P$.

Then, in patch embedding, the flattened patches are mapped to vectors with dimensions of $D$ by using a matrix $\mathbf{E} \in \mathbb{R}^{(P^2 C) \times D}$, and in position embedding, the position information $\mathbf{E}_\mathrm{pos} \in \mathbb{R}^{(N + 1) \times D}$ is embedded into each patch as

\begin{equation}
    \begin{aligned}
    z_0 &= [\boldsymbol{x}_\mathrm{class}; \boldsymbol{x}^1_\mathrm{p}\mathbf{E}; \boldsymbol{x}^2_\mathrm{p}\mathbf{E}; \cdots ; \boldsymbol{x}^N_\mathrm{p}\mathbf{E}] + \mathbf{E}_\mathrm{pos}, \\
    \mathbf{E}_\mathrm{pos} &= \begin{pmatrix}
     \left (\boldsymbol{e}^0_{\mathrm{pos}} \right )^{\mathsf{T}} & \left (\boldsymbol{e}^1_{\mathrm{pos}} \right )^{\mathsf{T}} & \cdots & \left (\boldsymbol{e}^N_{\mathrm{pos}} \right )^{\mathsf{T}}
    \end{pmatrix}^{\mathsf{T}} \\
    \boldsymbol{e}^i_{\mathrm{pos}} & \in \mathbb{R}^{D}, i=0,1,...,N
    \end{aligned}
\end{equation}
where $\boldsymbol{x}_\mathrm{class}$ is the classification token which is the input to MLP (see Fig. 2), $\boldsymbol{e}^0_{\mathrm{pos}}$ is the information of the classification token.

The proposed model transformation is carried out in accordance with the above relation.

\subsubsection{Position Embedding Transformed with Key}
Position embedding is an operation that embeds position information into  classification token and  each patch as in Eq. (7).
Let us define a matrix  $\mathbf{\hat{E}}_\mathrm{pos}$  consisting of position information  of each patch  as

\begin{equation}
    \begin{aligned}
    \mathbf{\hat{E}}_\mathrm{pos} &=
    \begin{pmatrix}
     \left (\boldsymbol{e}^1_{\mathrm{pos}} \right )^{\mathsf{T}} & \left (\boldsymbol{e}^2_{\mathrm{pos}} \right )^{\mathsf{T}} & \cdots & \left (\boldsymbol{e}^N_{\mathrm{pos}} \right )^{\mathsf{T}}
    \end{pmatrix}^{\mathsf{T}}, \\
    \boldsymbol{e}^i_{\mathrm{pos}} & \in \mathbb{R}^{D}, i=1,2,...,N.
    \end{aligned}
\end{equation}

Note that $\mathbf{\hat{E}}_\mathrm{pos}$ does not contain information about the classification token.

The permutation of rows in  $\mathbf{\hat{E}}_\mathrm{pos}$  corresponds to the block permutation in image transformation.
Therefore,  $\mathbf{\hat{E}}_\mathrm{pos}$  can be permuted using key $K_1$ used for block permutation, and the model can be encrypted.
The transformed model offers a high accuracy for only test images transformed by block permutation with $K_1$.
By defining a permutation matrix $\mathbf{E}_1$ with key $K_1$, the transformation from  $\mathbf{\hat{E}}_\mathrm{pos}$  to  $\mathbf{\hat{E}}'_\mathrm{pos}$  can be given as follows.

\begin{equation}
    \begin{aligned}
    \mathbf{\hat{E}}'_\mathrm{pos} &= \mathbf{E}_1 \mathbf{\hat{E}}_\mathrm{pos} \\
    &= \begin{pmatrix}
     \left (\boldsymbol{e'}^1_{\mathrm{pos}} \right )^{\mathsf{T}} & \cdots & \left (\boldsymbol{e'}^N_{\mathrm{pos}} \right )^{\mathsf{T}}
    \end{pmatrix}^{\mathsf{T}}, \\
    \boldsymbol{e'}^i_{\mathrm{pos}} & \in \{ \boldsymbol{e}^1_{\mathrm{pos}}, \cdots, \boldsymbol{e}^N_{\mathrm{pos}} \}
    \end{aligned}
\end{equation}
Therefore, from Eq. (7), the transformed matrix $\mathbf{E}'_\mathrm{pos}$ is given by
\begin{equation}
    \mathbf{E}'_\mathrm{pos} =
    \begin{pmatrix}
     \left (\boldsymbol{e}^0_{\mathrm{pos}} \right )^{\mathsf{T}} & \left (\boldsymbol{e'}^1_{\mathrm{pos}} \right )^{\mathsf{T}} & \cdots & \left (\boldsymbol{e'}^N_{\mathrm{pos}} \right )^{\mathsf{T}}
    \end{pmatrix}^{\mathsf{T}}.
\end{equation}

\subsubsection{Patch Embedding Transformed with Key}
In patch embedding, flattened patches are mapped to vectors with a dimension of $D$ as in Eq. (7).
When the patch size of ViT is equal to the block size of image transformation,  $P^2 C = p_\mathrm{b}$  is satisfied.
Therefore, the permutation of rows in $\mathbf{E}$ corresponds to pixel shuffling, so the model can be encrypted with key $K_2$ used for pixel shuffling.
The accuracy of the transformed model is high only when test images are encrypted by using pixel shuffling with key $K_2$.
As well as the transformation of $\mathbf{E}_{\mathrm{pos}}$, a permutation matrix $\mathbf{E}_2$ is defined with key $K_2$, and the transformation from matrix $\mathbf{E}$ to $\mathbf{E'}$ is shown as follows.

\begin{equation}
    \mathbf{E'} = \mathbf{E}_2 \mathbf{E}
\end{equation}

In addition, as shown in Eqs. (4) and (5), bit flipping with normalization can be regarded as an operation that randomly inverses the positive/negative sign of a pixel value.
Therefore, we can encrypt a model by inverting the sign of the rows in matrix $\mathbf{E}$ with key $K_3$ used for bit flipping.
The transformed model offers a high accuracy only for test images transformed by bit flipping with key $K_3$.
Using key $K_3$ to generate the same vector $\boldsymbol{r}$ used in bit flipping, the transformation from $\mathbf{E}$ to $\mathbf{E'}$ can be expressed as follows.

\begin{equation}
    \mathbf{E}'(k, :) = \left \{
    \begin{aligned}
    & \mathbf{E}(k, :) && (r_k=0) \\
    & - \mathbf{E}(k, :) && (r_k=1)
    \end{aligned}
    \right .
\end{equation}

\noindent
where $\mathbf{E}(k, :)$ and $\mathbf{E}'(k, :)$ are $k$-th rows of matrices $\mathbf{E}$ and $\mathbf{E}'$.\par
Accordingly, the procedure of the proposed method can be summarized as follows.
\begin{description}
    \item[\textbf{Step 1:}] Prepare a key set $K = \{ K_1, K_2, K_3 \}$.
    \item[\textbf{Step 2:}] 
    Generate $\mathbf{E}_1$ with $K_1$, $\mathbf{E}_2$ with $K_2$, and $\boldsymbol{r}$ with $K_3$
    
    \item[\textbf{Step 3:}] Transform a model $V_\theta$ trained with plain images by using $\mathbf{E}_1$, $\mathbf{E}_2$, and $\boldsymbol{r}$ on the basis of Eqs. (9), (10), and (11) as
    \begin{equation}
        V'_\theta = t(V_\theta, \{ \mathbf{E}_1, \mathbf{E}_2, \boldsymbol{r} \}),
    \end{equation}
    where $t(V_\theta, \{ \mathbf{E}_1, \mathbf{E}_2, \boldsymbol{r} \})$ is the proposed model transformation algorithm, and $V'_\theta$ is a transformed model.
    \item[\textbf{Step 4:}] Transform test images with $K = \{ K_1, K_2, K_3 \}$.
\end{description}

\subsection{Properties of Proposed Scheme}
The proposed method has the following properties.
\begin{itemize}
    \item The model performs well only if test images are transformed with the same key as that used for transforming the model.
    \item The proposed scheme does not cause performance degradation, due to the relation
    \begin{equation}
        V'_\theta(x') = V_\theta(x).
    \end{equation}
    \item Model training and encryption are independent. Therefore, it is possible to easily update a key.
\end{itemize}

\section{Experiment and Discussion}
In an experiment, the effectiveness of the proposed scheme was shown in terms of image classification accuracy and model protection performance.

\subsection{Experiment Setup}
To confirm the effectiveness of the proposed method, we evaluated the accuracy of an image classification task on the CIFAR-10 dataset (with 10 classes).
The CIFAR-10 consists of 60,000 color images (dimension of $3 \times 32 \times 32$), where 50,000 images are for training, 10,000 for testing, and each class contains 6,000 images.
Images in the dataset were resized to $3 \times224 \times 224$ to input them to ViT, before applying the proposed encryption algorithm, where the block size was $16 \times 16$. \par
We used the PyTorch \cite{paszke2019pytorch} implementation of ViT and fine-tuned a ViT model with a patch size $P=16$, which was pre-trained on ImageNet-21k.
The ViT model was fine-tuned for $5000$ epochs.
The parameters of the stochastic gradient descent (SGD) optimizer were a momentum of 0.9 and a learning rate value of 0.03. \par
In addition, we used three conventional visual information protection methods (Tanaka's method \cite{tanaka2018learnable}, the pixel-based encryption method \cite{sirichotedumrong2019privacy}, and the GAN-based transformation method \cite{sirichotedumrong2021a}) to compare them with the proposed method.
ResNet-20 was used to validate the effectiveness of the conventional method with reference to \cite{ito2021image}.
The CIFAR-10 was also used for training networks, and the networks were trained for $200$ epochs by using stochastic gradient descent (SGD) with a weight decay of $0.0005$ and a momentum of $0.9$. The learning rate was initially set to $0.1$, and it was multiplied by $0.2$ at $60$, $120$, and $160$ epochs.
The batch size was $128$.

\subsection{Image Classification}
First, we evaluated the proposed and conventional methods in terms of the accuracy of image classification under the use of ViT and ResNet-20.
As shown in Table 1, the performance of all conventional methods was degraded compared with the baselines.
In contrast, the proposed method did not degrade the performance at all. 

\begin{table}[t]
    \centering
    \caption{Comparison with conventional methods in terms of classification accuracy}
    \begin{tabular}{c|c|c}
    \hline
    Model & Method & Accuracy \\
    \hline
    \hline
    ViT & Baseline & 99.03 \\
    & Proposed & \textbf{99.03} \\
    \hline
    ResNet-20 \cite{ito2021image} & Baseline & 91.55 \\
    & Tanaka \cite{tanaka2018learnable} & 87.02 \\
    & Pixel-based \cite{sirichotedumrong2019privacy} & 86.66 \\
    & GAN-based \cite{sirichotedumrong2021a} & 82.55 \\
    \hline
    \end{tabular}
    \label{table1}
\end{table}

\subsection{Model Protection}
Next, we validated whether the proposed method has the ability to protect models.
Table 2 shows the accuracy of image classification when encrypted or plain images were input to the encrypted model.

The encrypted model performed well for test images with the correct key, but its accuracy was not high when using plain test images.
The CIFAR-10 dataset consists of ten classes, so $9.06$ is almost the same accuracy as that when test images are randomly classified.

\par

\begin{table}[t]
    \centering
    \caption{Robustness against use of plain images}
    \begin{tabular}{c|cc}
    \hline
    & \multicolumn{2}{c}{Test Image} \\
    \cline{2-3}
    Model & Plain & Proposed \\
    \hline
    Baseline & \textbf{99.03} & - \\
    Proposed & 9.06 & \textbf{99.03} \\
    \hline
    \end{tabular}
    \label{tab:my_label}
\end{table}

Next, we confirmed the performance of images encrypted with a different key from that used in the model encryption.
We prepared 100 random keys, and test images encrypted with the keys were input to the encrypted model.
From the box plot in Fig. 5, the accuracy of the models was not high under the use of the wrong keys.
Accordingly, the proposed scheme was confirmed to be robust against a random key attack. \par
The use of a large key space enhances robustness against various attacks in general.
In this experiment, the key space of block permutation, pixel shuffling, and bit flipping ($O_\mathsf{P}$, $O_\mathsf{S}$, and $O_\mathsf{F}$) is given by

\begin{equation}
    \begin{aligned}
    O_{\mathrm{P}} &= (W_{\mathrm{b}} H_{\mathrm{b}})! \\
    &= (WH/M^2)! = 196!, 
    \end{aligned}
\end{equation}

\begin{equation}
    \begin{aligned}
    O_{\mathrm{S}} &= p_{\mathrm{b}}! \\
    &= (M^2C)! = 768!, 
    \end{aligned}
\end{equation}

\noindent
and

\begin{equation}
    \begin{aligned}
    O_{\mathrm{F}} & = \frac{p_{\mathrm{b}}!}{(p_{\mathrm{b}}/2)! \cdot (p_{\mathrm{b}}/2)!} \\
    & = \frac{(M^2C)!}{(M^2C/2)! \cdot (M^2C/2)!} = \frac{768!}{384! \cdot 384!}.
    \end{aligned}
\end{equation}

\noindent
Therefore, the key space of the proposed method is represented as follows.

\begin{equation}
    O = O_{\mathrm{P}} \times O_{\mathrm{S}} \times O_{\mathrm{F}} \simeq 2^{8237}
\end{equation}

\noindent

Typical cipher systems are recommended to have $2^{128}$ as a key space as in \cite{schneier1995applied}.
Accordingly, the key space $O$ is sufficiently large, so it is difficult to find the correct key by random key estimation.

\begin{figure}[t]
    \centering
    \includegraphics[scale=0.5]{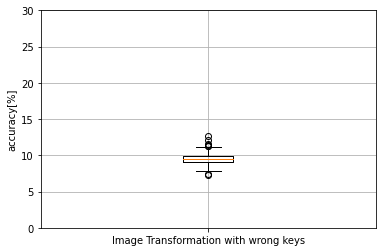}
    \caption{Evaluating robustness against random key attack. Boxes span from first to third quartile, referred to as $Q1$ and $Q3$, and whiskers show maximum and minimum values in range of $[Q1 - 1.5(Q3 - Q1), Q3 + 1.5(Q3 - Q1)]$. Band inside box indicates median. Outliers are indicated as dots.}
    \label{fig5}
\end{figure}

\section{Conclusion}
In this paper, we proposed the combined use of an image transformation method with a secret key and ViT models transformed with the key.
The proposed scheme enables us not only to use visually protected images but to also maintain the same classification accuracy as that of models trained with plain images.
In addition, in an experiment, the proposed scheme was demonstrated to outperform state-of-the-art methods with perceptually encrypted images in terms of classification accuracy, and it was also verified to be effective in model protection.

\section*{Acknowledgments}
This study was partially supported by JSPS KAKENHI (Grant Number $\mathrm{JP}\mathrm{21} \mathrm{H}\mathrm{01327}$) and JST CREST (Grant Number $\mathrm{JPMJCR} \mathrm{20} \mathrm{D} \mathrm{3}$).

\bibliographystyle{ieicetr}
\bibliography{output}

\begin{thebibliography}{10}
\providecommand{\url}[1]{#1}
\csname url@samestyle\endcsname
\providecommand{\newblock}{\relax}
\providecommand{\bibinfo}[2]{#2}
\providecommand{\BIBentrySTDinterwordspacing}{\spaceskip=0pt\relax}
\providecommand{\BIBentryALTinterwordstretchfactor}{4}
\providecommand{\BIBentryALTinterwordspacing}{\spaceskip=\fontdimen2\font plus
\BIBentryALTinterwordstretchfactor\fontdimen3\font minus
  \fontdimen4\font\relax}
\providecommand{\BIBforeignlanguage}[2]{{%
\expandafter\ifx\csname l@#1\endcsname\relax
\typeout{** WARNING: IEEEtran.bst: No hyphenation pattern has been}%
\typeout{** loaded for the language `#1'. Using the pattern for}%
\typeout{** the default language instead.}%
\else
\language=\csname l@#1\endcsname
\fi
#2}}
\providecommand{\BIBdecl}{\relax}
\BIBdecl

\bibitem{lecun2015deep}
Y.~LeCun, Y.~Bengio, and G.~Hinton, ``Deep learning,'' \emph{nature}, vol. 521,
  no. 7553, p. 436, 2015.

\bibitem{liu2019recent}
X.~Liu, Z.~Deng, and Y.~Yang, ``Recent progress in semantic image
  segmentation,'' \emph{Artif. Intell. Rev.}, vol.~52, no.~2, pp. 1089--1106,
  2019.

\bibitem{huang2014survey}
C.-T. Huang, L.~Huang, Z.~Qin, H.~Yuan, L.~Zhou, V.~Varadharajan, and C.-C.~J.
  Kuo, ``Survey on securing data storage in the cloud,'' \emph{APSIPA
  Transactions on Signal and Information Processing}, vol.~3, p.~e7, 2014.

\bibitem{moo2013p3}
M.-R. Ra, R.~Govindan, and A.~Ortega, ``P3: Toward {Privacy-Preserving} photo
  sharing,'' in \emph{10th USENIX Symposium on Networked Systems Design and
  Implementation (NSDI 13)}.\hskip 1em plus 0.5em minus 0.4em\relax Lombard,
  IL: USENIX Association, Apr. 2013, pp. 515--528.

\bibitem{lagendijk2013encrypted}
R.~Lagendijk, Z.~Erkin, and M.~Barni, ``Encrypted signal processing for privacy
  protection: Conveying the utility of homomorphic encryption and multiparty
  computation,'' \emph{IEEE Signal Processing Magazine}, vol.~30, no.~1, pp.
  82--105, 2013.

\bibitem{fredrikson2015model}
M.~Fredrikson, S.~Jha, and T.~Ristenpart, ``Model inversion attacks that
  exploit confidence information and basic countermeasures,'' in
  \emph{Proceedings of the 22nd ACM SIGSAC Conference on Computer and
  Communications Security}, ser. CCS '15.\hskip 1em plus 0.5em minus
  0.4em\relax New York, NY, USA: Association for Computing Machinery, 2015, pp.
  1322--1333.

\bibitem{shokri2017membership}
R.~Shokri, M.~Stronati, C.~Song, and V.~Shmatikov, ``Membership inference
  attacks against machine learning models,'' in \emph{2017 IEEE symposium on
  security and privacy (SP)}.\hskip 1em plus 0.5em minus 0.4em\relax IEEE,
  2017, pp. 3--18.

\bibitem{christian2014intriguing}
C.~Szegedy, W.~Zaremba, I.~Sutskever, J.~Bruna, D.~Erhan, I.~J. Goodfellow, and
  R.~Fergus, ``Intriguing properties of neural networks,'' in \emph{2nd
  International Conference on Learning Representations, {ICLR} 2014, Banff, AB,
  Canada, April 14-16, 2014, Conference Track Proceedings}, Y.~Bengio and
  Y.~LeCun, Eds., 2014.

\bibitem{siva2020adversarial}
R.~S. Siva~Kumar, M.~Nyström, J.~Lambert, A.~Marshall, M.~Goertzel,
  A.~Comissoneru, M.~Swann, and S.~Xia, ``Adversarial machine learning-industry
  perspectives,'' in \emph{2020 IEEE Security and Privacy Workshops (SPW)},
  2020, pp. 69--75.

\bibitem{kiya2022overview}
\BIBentryALTinterwordspacing
H.~Kiya, A.~P.~M. Maung, Y.~Kinoshita, S.~Imaizumi, and S.~Shiota, ``An
  overview of compressible and learnable image transformation with secret key
  and its applications,'' \emph{APSIPA Transactions on Signal and Information
  Processing}, vol.~11, no. 1, e11, 2022. [Online]. Available:
  \url{http://dx.doi.org/10.1561/116.00000048}
\BIBentrySTDinterwordspacing

\bibitem{aprilpyone2021block}
M.~Aprilpyone and H.~Kiya, ``Block-wise image transformation with secret key
  for adversarially robust defense,'' \emph{IEEE Transactions on Information
  Forensics and Security}, vol.~16, pp. 2709--2723, 2021.

\bibitem{maungmaung2021a}
------, ``A protection method of trained cnn model with a secret key from
  unauthorized access,'' \emph{APSIPA Transactions on Signal and Information
  Processing}, vol.~10, p. e10, 2021.

\bibitem{maung2020encryption}
------, ``Encryption inspired adversarial defense for visual classification,''
  in \emph{2020 IEEE International Conference on Image Processing (ICIP)},
  2020, pp. 1681--1685.

\bibitem{maung2021ensemble}
------, ``Ensemble of key-based models: Defense against black-box adversarial
  attacks,'' in \emph{2021 IEEE 10th Global Conference on Consumer Electronics
  (GCCE)}, 2021, pp. 95--98.

\bibitem{april2022privacy}
------, ``Privacy-preserving image classification using isotropic network,''
  \emph{IEEE MultiMedia}, vol.~29, no.~2, pp. 23--33, 2022.

\bibitem{qi2022privacy}
\BIBentryALTinterwordspacing
Z.~Qi, M.~Aprilpyone, Y.~Kinoshita, and H.~Kiya, ``Privacy-preserving image
  classification using vision transformer,'' \emph{arXiv preprint
  arXiv:1804.00750}, 2022. [Online]. Available:
  \url{https://arxiv.org/abs/2205.12041}
\BIBentrySTDinterwordspacing

\bibitem{Alexey2021an}
A.~Dosovitskiy, L.~Beyer, A.~Kolesnikov, D.~Weissenborn, X.~Zhai,
  T.~Unterthiner, M.~Dehghani, M.~Minderer, G.~Heigold, S.~Gelly, J.~Uszkoreit,
  and N.~Houlsby, ``An image is worth 16x16 words: Transformers for image
  recognition at scale,'' in \emph{International Conference on Learning
  Representations}, 2021.

\bibitem{chuman2019encryption}
T.~Chuman, W.~Sirichotedumrong, and H.~Kiya, ``Encryption-then-compression
  systems using grayscale-based image encryption for jpeg images,'' \emph{IEEE
  Transactions on Information Forensics and Security}, vol.~14, no.~6, pp.
  1515--1525, 2019.

\bibitem{chuman2017security}
T.~Chuman, K.~Kurihara, and H.~Kiya, ``On the security of block
  scrambling-based etc systems against jigsaw puzzle solver attacks,'' in
  \emph{2017 IEEE International Conference on Acoustics, Speech and Signal
  Processing (ICASSP)}, 2017, pp. 2157--2161.

\bibitem{zhou2014designing}
J.~Zhou, X.~Liu, O.~C. Au, and Y.~Y. Tang, ``Designing an efficient image
  encryption-then-compression system via prediction error clustering and random
  permutation,'' \emph{IEEE Transactions on Information Forensics and
  Security}, vol.~9, no.~1, pp. 39--50, 2014.

\bibitem{ghonge2014a}
M.~Ghonge and K.~Nimbokar, ``A survey based on designing an efficient image
  encryption-then-compression system,'' \emph{International Journal of Computer
  Applications}, p. 8887, 2014.

\bibitem{liu2018ecg}
T.~Y. Liu, K.~J. Lin, and H.~C. Wu, ``Ecg data encryption then compression
  using singular value decomposition,'' \emph{IEEE Journal of Biomedical and
  Health Informatics}, vol.~22, no.~3, pp. 707--713, 2018.

\bibitem{liu2010efficient}
W.~Liu, W.~Zeng, L.~Dong, and Q.~Yao, ``Efficient compression of encrypted
  grayscale images,'' \emph{IEEE Transactions on Image Processing}, vol.~19,
  no.~4, pp. 1097--1102, 2010.

\bibitem{hu2014a}
R.~Hu, X.~Li, and B.~Yang, ``A new lossy compression scheme for encrypted
  gray-scale images,'' in \emph{2014 IEEE International Conference on
  Acoustics, Speech and Signal Processing (ICASSP)}, 2014, pp. 7387--7390.

\bibitem{johnson2004on}
M.~Johnson, P.~Ishwar, V.~Prabhakaran, D.~Schonberg, and K.~Ramchandran, ``On
  compressing encrypted data,'' \emph{IEEE Transactions on Signal Processing},
  vol.~52, no.~10, pp. 2992--3006, 2004.

\bibitem{methaq2016an}
M.~Gaata and F.~F. Hantoosh, ``An efficient image encryption technique using
  chaotic logistic map and rc4 stream cipher,'' \emph{International Journal of
  Modern Trends in Engineering and Research}, vol.~3, pp. 213--218, 2016.

\bibitem{watanabe2015an}
O.~Watanabe, A.~Uchida, T.~Fukuhara, and H.~Kiya, ``An
  encryption-then-compression system for jpeg 2000 standard,'' \emph{2015 IEEE
  International Conference on Acoustics, Speech and Signal Processing
  (ICASSP)}, pp. 1226--1230, 2015.

\bibitem{Shoko2020a}
S.~Imaizumi, Y.~Izawa, R.~Hirasawa, and H.~Kiya, ``A reversible data hiding
  method in compressible encrypted images,'' \emph{IEICE Transactions on
  Fundamentals of Electronics, Communications and Computer Sciences}, vol.
  E103.A, no.~12, pp. 1579--1588, 2020.

\bibitem{iida2020privacy}
K.~Iida and H.~Kiya, ``Privacy-preserving content-based image retrieval using
  compressible encrypted images,'' \emph{IEEE Access}, vol.~8, pp.
  200\,038--200\,050, 2020.

\bibitem{iida2019an}
------, ``An image identification scheme of encrypted jpeg images for
  privacy-preserving photo sharing services,'' \emph{2019 IEEE International
  Conference on Image Processing (ICIP)}, pp. 4564--4568, 2019.

\bibitem{liu2022block}
\BIBentryALTinterwordspacing
S.~Liu, Z.~Lian, S.~Gu, and L.~Xiao, ``Block shuffling learning for deepfake
  detection,'' \emph{arXiv preprint arXiv:2202.02819}, 2022. [Online].
  Available: \url{https://arxiv.org/abs/2202.02819}
\BIBentrySTDinterwordspacing

\bibitem{kawamura2020aprivacy}
A.~Kawamura, Y.~Kinoshita, T.~Nakachi, S.~Shiota, and H.~Kiya, ``A
  privacy-preserving machine learning scheme using etc images,'' \emph{IEICE
  Transactions on Fundamentals of Electronics, Communications and Computer
  Sciences}, vol. E103.A, no.~12, pp. 1571--1578, 2020.

\bibitem{bandoh2020distributed}
Y.~Bandoh, T.~Nakachi, and H.~Kiya, ``Distributed secure sparse modeling based
  on random unitary transform,'' \emph{IEEE Access}, vol.~8, pp.
  211\,762--211\,772, 2020.

\bibitem{takayuki2020secure}
T.~Nakachi, Y.~Bandoh, and H.~Kiya, ``Secure overcomplete dictionary learning
  for sparse representation,'' \emph{IEICE Transactions on Information and
  Systems}, vol. E103.D, no.~1, pp. 50--58, 2020.

\bibitem{nakachi2020privacy}
T.~Nakachi, Y.~Wang, and H.~Kiya, ``Privacy-preserving pattern recognition
  using encrypted sparse representations in l0 norm minimization,'' in
  \emph{ICASSP 2020 - 2020 IEEE International Conference on Acoustics, Speech
  and Signal Processing (ICASSP)}, 2020, pp. 2697--2701.

\bibitem{ibuki2016unitary}
I.~Nakamura, Y.~Tonomura, and H.~Kiya, ``Unitary transform-based template
  protection and its application to {$l^{2}$}-norm minimization problems,''
  \emph{IEICE Transactions on Information and Systems}, vol. E99.D, no.~1, pp.
  60--68, 2016.

\bibitem{maekawa2019privacy}
T.~Maekawa, A.~Kwamura, T.~Nakachi, and H.~Kiya, ``Privacy-preserving support
  vector machine computing using random unitary transformation,'' \emph{IEICE
  Transactions on Fundamentals of Electronics, Communications and Computer
  Sciences}, vol. E102.A, no.~12, pp. 1849--1855, 2019.

\bibitem{chen2018protect}
M.~Chen and M.~Wu, ``Protect your deep neural networks from piracy,'' in
  \emph{2018 IEEE International Workshop on Information Forensics and Security
  (WIFS)}, 2018, pp. 1--7.

\bibitem{chen2019deepattest}
H.~Chen, C.~Fu, B.~D. Rouhani, J.~Zhao, and F.~Koushanfar, ``Deepattest: An
  end-to-end attestation framework for deep neural networks,'' in \emph{2019
  ACM/IEEE 46th Annual International Symposium on Computer Architecture
  (ISCA)}, 2019, pp. 487--498.

\bibitem{uchida2017embedding}
Y.~Uchida, Y.~Nagai, S.~Sakazawa, and S.~Satoh, ``Embedding watermarks into
  deep neural networks,'' in \emph{Proceedings of the 2017 ACM on International
  Conference on Multimedia Retrieval}, ser. ICMR '17.\hskip 1em plus 0.5em
  minus 0.4em\relax Association for Computing Machinery, 2017, pp. 269--277.

\bibitem{chen2019deepmarks}
H.~Chen, B.~D. Rouhani, C.~Fu, J.~Zhao, and F.~Koushanfar, ``Deepmarks: A
  secure fingerprinting framework for digital rights management of deep
  learning models,'' in \emph{Proceedings of the 2019 on International
  Conference on Multimedia Retrieval}, ser. ICMR '19.\hskip 1em plus 0.5em
  minus 0.4em\relax Association for Computing Machinery, 2019, p. 105–113.

\bibitem{rouhani2018deepsigns}
\BIBentryALTinterwordspacing
B.~D. Rouhani, H.~Chen, and F.~Koushanfar, ``Deepsigns: A generic watermarking
  framework for ip protection of deep learning models,'' \emph{arXiv preprint
  arXiv:1804.00750}, 2018. [Online]. Available:
  \url{https://arxiv.org/abs/1804.00750}
\BIBentrySTDinterwordspacing

\bibitem{fan2021deepip}
L.~Fan, K.~W. Ng, C.~S. Chan, and Q.~Yang, ``Deepip: Deep neural network
  intellectual property protection with passports,'' \emph{IEEE Transactions on
  Pattern Analysis and Machine Intelligence}, 2021.

\bibitem{adi2018turning}
Y.~Adi, C.~Baum, M.~Cisse, B.~Pinkas, and J.~Keshet, ``Turning your weakness
  into a strength: Watermarking deep neural networks by backdooring,'' in
  \emph{Proceedings of the 27th USENIX Conference on Security Symposium}, ser.
  SEC'18.\hskip 1em plus 0.5em minus 0.4em\relax USENIX Association, 2018, pp.
  1615--1631.

\bibitem{zhang2018protecting}
J.~Zhang, Z.~Gu, J.~Jang, H.~Wu, M.~P. Stoecklin, H.~Huang, and I.~Molloy,
  ``Protecting intellectual property of deep neural networks with
  watermarking,'' in \emph{Proceedings of the 2018 on Asia Conference on
  Computer and Communications Security}, ser. ASIACCS '18.\hskip 1em plus 0.5em
  minus 0.4em\relax Association for Computing Machinery, 2018, pp. 159--172.

\bibitem{sakazawa2019visual}
S.~Sakazawa, E.~Myodo, K.~Tasaka, and H.~Yanagihara, ``Visual decoding of
  hidden watermark in trained deep neural network,'' in \emph{2019 IEEE
  Conference on Multimedia Information Processing and Retrieval (MIPR)}, 2019,
  pp. 371--374.

\bibitem{le2019adversarial}
E.~Le~Merrer, P.~P{\'e}rez, and G.~Tr{\'e}dan, ``{Adversarial frontier
  stitching for remote neural network watermarking},'' \emph{{Neural Computing
  and Applications}}, vol.~32, no.~13, pp. 9233--9244, 2019.

\bibitem{devlin2019bert}
J.~Devlin, M.-W. Chang, K.~Lee, and K.~Toutanova, ``{BERT}: Pre-training of
  deep bidirectional transformers for language understanding,'' in
  \emph{Proceedings of the 2019 Conference of the North {A}merican Chapter of
  the Association for Computational Linguistics: Human Language Technologies,
  Volume 1 (Long and Short Papers)}.\hskip 1em plus 0.5em minus 0.4em\relax
  Association for Computational Linguistics, Jun. 2019, pp. 4171--4186.

\bibitem{ilya2021mlpmixer}
I.~Tolstikhin, N.~Houlsby, A.~Kolesnikov, L.~Beyer, X.~Zhai, T.~Unterthiner,
  J.~Yung, A.~P. Steiner, D.~Keysers, J.~Uszkoreit, M.~Lucic, and
  A.~Dosovitskiy, ``{MLP}-mixer: An all-{MLP} architecture for vision,'' in
  \emph{Advances in Neural Information Processing Systems}, A.~Beygelzimer,
  Y.~Dauphin, P.~Liang, and J.~W. Vaughan, Eds., 2021.

\bibitem{hugo2021resmlp}
H.~Touvron, P.~B. andMathilde Caron, M.~Cord, A.~El{-}Nouby, E.~Grave,
  A.~Joulin, G.~Synnaeve, J.~Verbeek, and H.~J{\'{e}}gou, ``Resmlp: Feedforward
  networks for image classification with data-efficient training,''
  \emph{CoRR}, vol. abs/2105.03404, 2021.

\bibitem{chen2022cyclemlp}
S.~Chen, E.~Xie, C.~GE, R.~Chen, D.~Liang, and P.~Luo, ``Cycle{MLP}: A
  {MLP}-like architecture for dense prediction,'' in \emph{International
  Conference on Learning Representations}, 2022.

\bibitem{liu2021pay}
H.~Liu, Z.~Dai, D.~So, and Q.~V. Le, ``Pay attention to {MLP}s,'' in
  \emph{Advances in Neural Information Processing Systems}, A.~Beygelzimer,
  Y.~Dauphin, P.~Liang, and J.~W. Vaughan, Eds., 2021.

\bibitem{Hou2022VisionPA}
Q.~Hou, Z.~Jiang, L.~Yuan, M.-M. Cheng, S.~Yan, and J.~Feng, ``Vision
  permutator: A permutable mlp-like architecture for visual recognition,''
  \emph{IEEE transactions on pattern analysis and machine intelligence},
  vol.~PP, 2022.

\bibitem{trockman2022patches}
\BIBentryALTinterwordspacing
A.~Trockman and J.~Z. Kolter, ``Patches are all you need?'' \emph{arXiv
  preprint arXiv:2201.09792}, 2022. [Online]. Available:
  \url{https://arxiv.org/abs/2201.09792}
\BIBentrySTDinterwordspacing

\bibitem{paszke2019pytorch}
A.~Paszke, S.~Gross, F.~Massa, A.~Lerer, J.~Bradbury, G.~Chanan, T.~Killeen,
  Z.~Lin, N.~Gimelshein, L.~Antiga, A.~Desmaison, A.~Kopf, E.~Yang, Z.~DeVito,
  M.~Raison, A.~Tejani, S.~Chilamkurthy, B.~Steiner, L.~Fang, J.~Bai, and
  S.~Chintala, ``Pytorch: An imperative style, high-performance deep learning
  library,'' in \emph{Advances in Neural Information Processing Systems 32},
  H.~Wallach, H.~Larochelle, A.~Beygelzimer, F.~d\textquotesingle
  Alch\'{e}-Buc, E.~Fox, and R.~Garnett, Eds.\hskip 1em plus 0.5em minus
  0.4em\relax Curran Associates, Inc., 2019, pp. 8024--8035.

\bibitem{tanaka2018learnable}
M.~Tanaka, ``Learnable image encryption,'' in \emph{2018 IEEE International
  Conference on Consumer Electronics-Taiwan (ICCE-TW)}, 2018, pp. 1--2.

\bibitem{sirichotedumrong2019privacy}
W.~Sirichotedumrong, T.~Maekawa, Y.~Kinoshita, and H.~Kiya,
  ``Privacy-preserving deep neural networks with pixel-based image encryption
  considering data augmentation in the encrypted domain,'' in \emph{2019 IEEE
  International Conference on Image Processing (ICIP)}, 2019, pp. 674--678.

\bibitem{sirichotedumrong2021a}
W.~Sirichotedumrong and H.~Kiya, ``A gan-based image transformation scheme for
  privacy-preserving deep neural networks,'' in \emph{2020 28th European Signal
  Processing Conference (EUSIPCO)}, 2021, pp. 745--749.

\bibitem{ito2021image}
H.~Ito, Y.~Kinoshita, M.~Aprilpyone, and H.~Kiya, ``Image to perturbation: An
  image transformation network for generating visually protected images for
  privacy-preserving deep neural networks,'' \emph{IEEE Access}, vol.~9, pp.
  64\,629--64\,638, 2021.

\bibitem{schneier1995applied}
B.~Schneier and P.~Sutherland, \emph{Applied Cryptography: Protocols,
  Algorithms, and Source Code in C}, 2nd~ed.\hskip 1em plus 0.5em minus
  0.4em\relax USA: John Wiley and Sons, Inc., 1995.

\end{thebibliography}

\profile[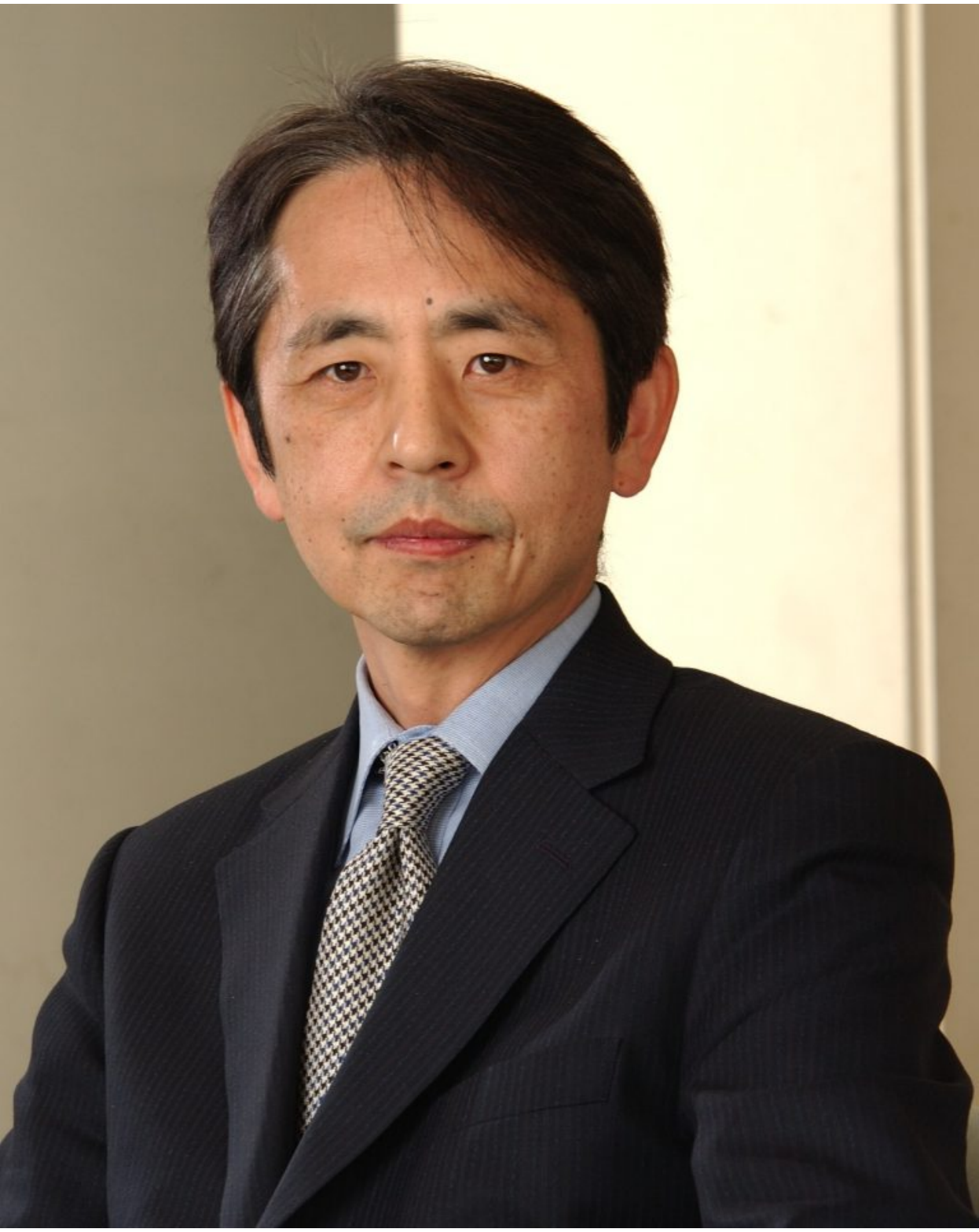]{Hitoshi Kiya}{ 
received his B.E and M.E. degrees from the Nagaoka University of
Technology in 1980 and 1982, respectively, and his Dr. Eng. degree from Tokyo Metropolitan University in 1987.
In 1982, he joined Tokyo Metropolitan University, where he became a Full Professor in 2000.
From 1995 to 1996, he attended the University of Sydney, Australia as a Visiting Fellow.
He is a Fellow of IEEE, IEICE, AAIA and ITE.
He served as President of APSIPA from 2019 to 2020 and as Regional Director-at-Large for Region 10 of the IEEE Signal Processing Society from 2016 to 2017.
He was also President of the IEICE Engineering Sciences Society from 2011 to 2012, and he served there as Editor in-Chief for IEICE Society Magazine and Society Publications.
He has been an Editorial Board Member of eight journals, including IEEE Trans. on Signal Processing, Image Processing, and Information  Forensics and Security, Chair of two technical committees, and Member of nine technical committees including the APSIPA Image, Video, and Multimedia Technical Committee (TC) and IEEE Information Forensics and Security TC.
He has organized a lot of international conferences in such roles as TPC Chair of IEEE ICASSP 2012 and as General Co-Chair of IEEE ISCAS 2019.
He has received numerous awards, including 12 best paper awards.
}

\profile[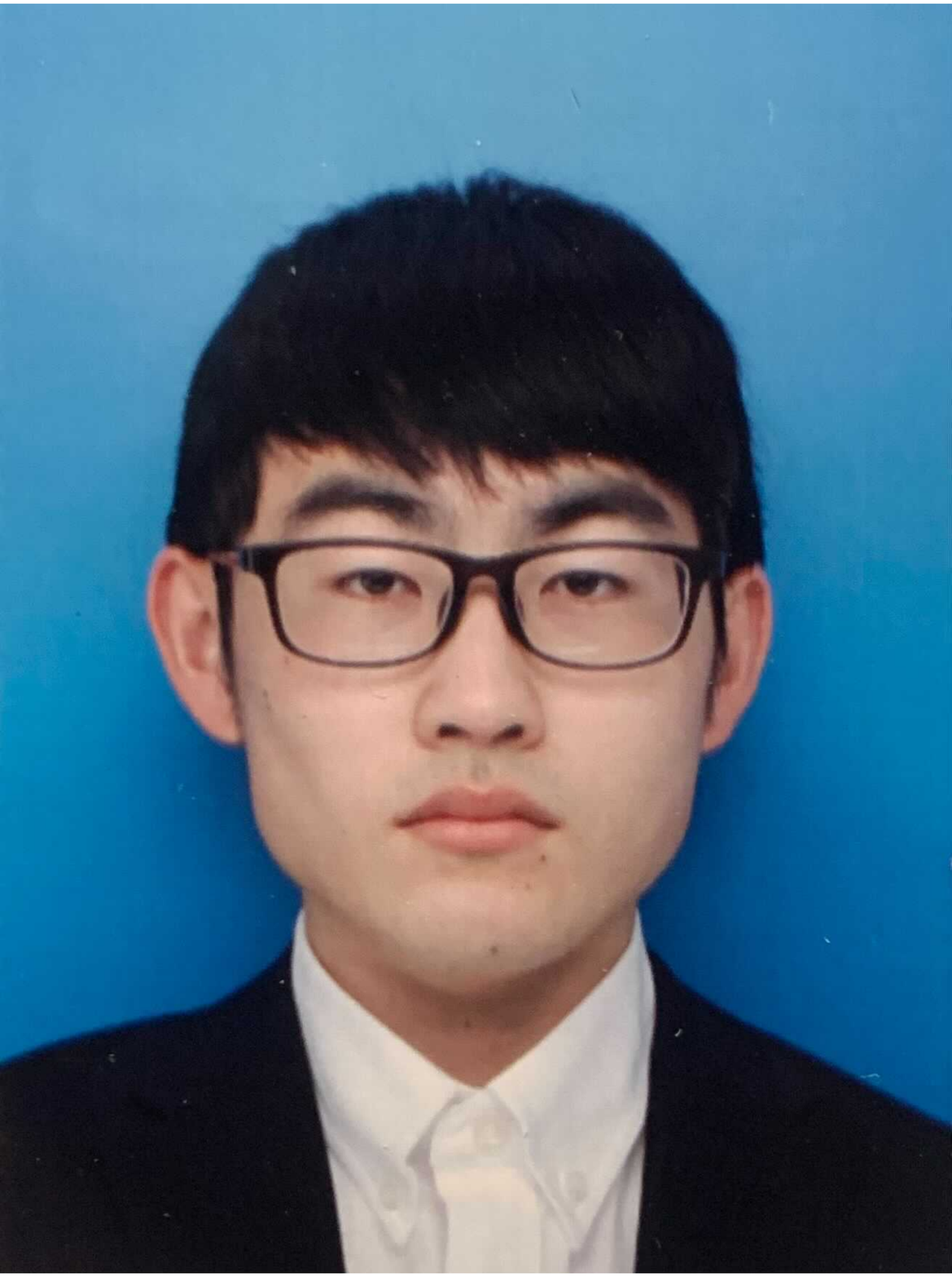]{Ryota Iijima}{received his B.C.S degree from Tokyo Metropolitan University, Japan in 2022.
Since 2022, he has been a Master course student at Tokyo Metropolitan University. 
His research interests include deep neural networks and their protection.}

\profile[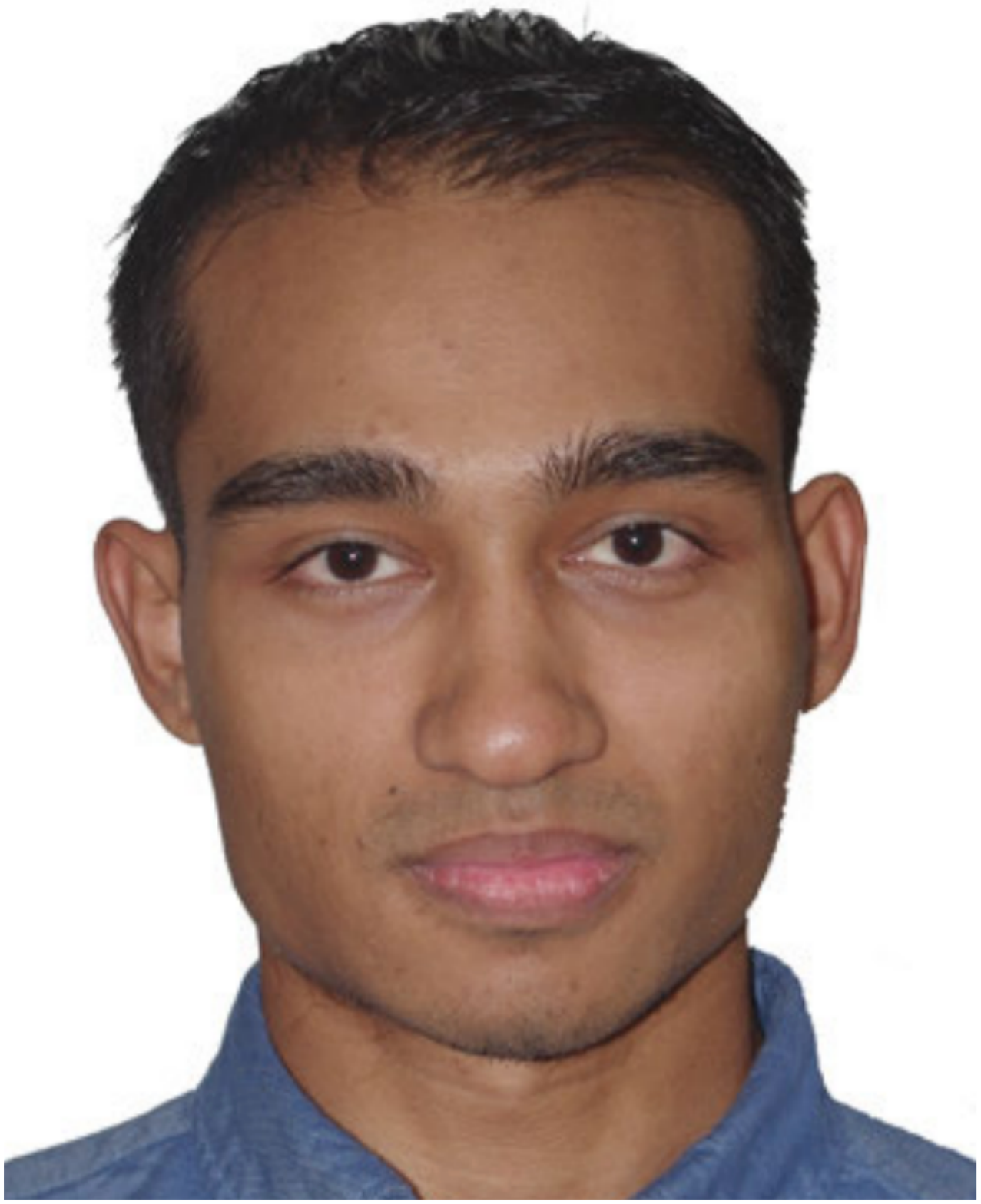]{MaungMaung AprilPyone}{received a BCS degree from the International Islamic University Malaysia in 2013 under the Albukhary Foundation Scholarship, MCS degree from the University of Malaya in 2018 under the International Graduate Research Assistantship Scheme, and Ph.D. degree from the Tokyo Metropolitan University in 2022 under the Tokyo Human Resources Fund for City Diplomacy Scholarship. He is currently working as a Project Assistant Professor in the Tokyo Metropolitan University and as a researcher in rinna Co. Ltd. He received an IEEE ICCE-TW Best Paper Award in 2016. His research interests are in the area of adversarial machine learning and information security. He is a member of IEEE.}

\profile[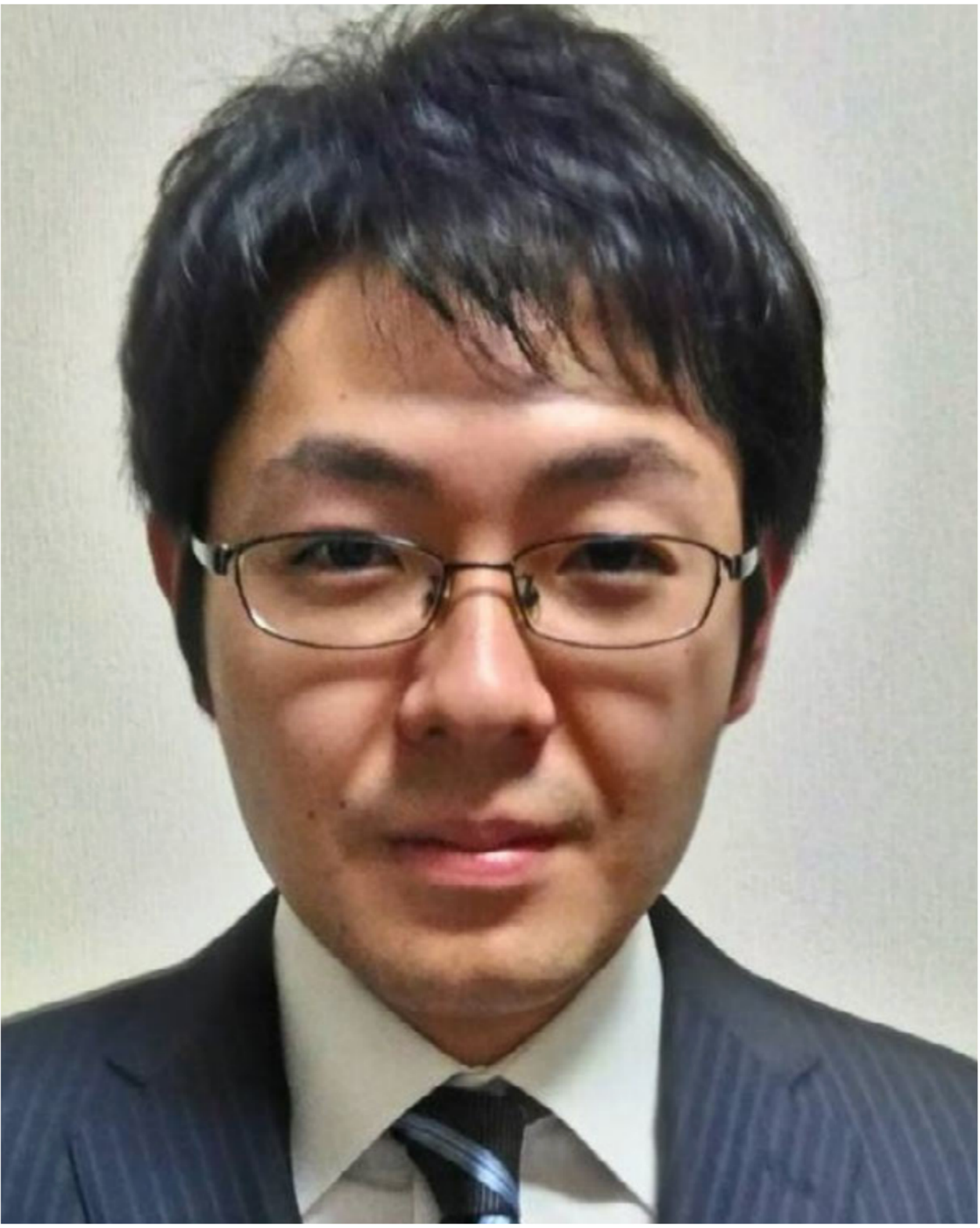]{Yuma Kinoshita}{received his B.Eng., M.Eng., and the Ph.D. degrees from Tokyo Metropolitan University, Japan, in 2016, 2018, and 2020 respectively. In April 2020, he started to work with Tokyo Metropolitan University, as a project assistant professor. He moved to Tokai University, Japan, as an associate professor/lecturer in April 2022. His research interests are in the area of signal processing, image processing, and machine learning. He is a Member of IEEE, APSIPA, IEICE, and ASJ. He received the IEEE ISPACS Best Paper Award, in 2016, the IEEE Signal Processing Society Japan Student Conference Paper Award, in 2018, the IEEE Signal Processing Society Tokyo Joint Chapter Student Award, in 2018, the IEEE GCCE Excellent Paper Award (Gold Prize), in 2019, and the IWAIT Best Paper Award, in 2020. He was a Registration Chair of DCASE2020 Workshop.}

\end{document}